\documentclass{article}

\usepackage{amsmath}
\usepackage{graphicx}
\graphicspath{ {figures/} }
\usepackage[final]{nips_2017_workshop}

\usepackage[utf8]{inputenc} 
\usepackage[T1]{fontenc}    
\usepackage{hyperref}       
\usepackage{url}            
\usepackage{booktabs}       
\usepackage{amsfonts}       
\usepackage{nicefrac}       
\usepackage{microtype}      

\title{Calibrated Boosting-Forest}

\author{
  Haozhen Wu\\
  Department of Computer Science \& Small Molecule Screening Facility\\
  University of Wisconsin Madison\\
  \texttt{hwu84@wisc.edu} \\
}

\begin{document}

\maketitle

\begin{abstract}
Excellent ranking power along with well calibrated probability estimates are needed in many classification tasks. In this paper, we introduce a technique, \textit{Calibrated Boosting-Forest}\footnote{Code published at https://github.com/haozhenWu/Calibrated-Boosting-Forest} that captures both. This novel technique is an ensemble of gradient boosting machines that can support both continuous and binary labels. While offering superior ranking power over any individual regression or classification model, Calibrated Boosting-Forest is able to preserve well calibrated posterior probabilities. Along with these benefits, we provide an alternative to the tedious step of tuning gradient boosting machines. We demonstrate that tuning Calibrated Boosting-Forest can be reduced to a simple hyper-parameter selection. We further establish that increasing this hyper-parameter improves the ranking performance under a diminishing return. We examine the effectiveness of Calibrated Boosting-Forest on ligand-based virtual screening where both continuous and binary labels are available and compare the performance of Calibrated Boosting-Forest with logistic regression, gradient boosting machine and deep learning. Calibrated Boosting-Forest achieved an approximately 48\% improvement compared to a state-of-art deep learning model. Moreover, it achieved around 95\% improvement on probability quality measurement compared to the best individual gradient boosting machine. Calibrated Boosting-Forest offers a benchmark demonstration that in the field of ligand-based virtual screening, deep learning is not the universally dominant machine learning model and good calibrated probabilities can better facilitate virtual screening process.
\end{abstract}

\section{Introduction}

Any increasing or decreasing score that represents the probability of each class is sufficient to produce a correct ranking of examples [9]. If the label is under a continuous distribution, typically, we either apply regression, or convert the label to a classification problem. Both techniques can be used to effectively rank. Current research in this area has been focused on improving the accuracy of an individual model. We propose an integrative technique here called Calibrated Boosting-Forest (CBF) that adopts both characteristics from regression and classification that achieves better ranking performance. In order to bridge these approaches, we exploit \textit{stacked generalization} [1].

Stacked generalization is a methodology that links multiple base learners to construct a sharpened predictor. Using the original features, multiple base learners are trained and used for predictions. The base learners' predictions are then used as the feature matrix to train a second layer model, a meta-learner. Using this multiple-stage modeling approach, stacked generalization is able to offer final scores that exhibit both low bias and low variance. Because of the dependence between the base learners and the layer2 model, in order to avoid over-fitting a necessary precaution is to train the meta-learner on a subset of training data that is disjoint from that of which the base learners were trained upon. An enhanced version of this process is to use a cross-validation procedure that iteratively predicts on a valid set and then concludes by reassembling the predictions to their original order. This procedure is advantageous because it ensures the number of records in the layer2 data is equivalent to that of the layer1 data.

We chose gradient boosting machines, or GBMs [3] as fundamental models of CBF due to their excellent performance on ranking tasks, especially while using decision trees as base learners. GBMs are a way to consecutively fit new models to correct previous errors and provide a more accurate approximation of the response variable. During each iteration, a new weak learner is trained based on the error of the overall ensemble up to that point. GBMs primarily require a decision to be made on two components, the cost function and the weak learner. In other words, one has to specify what function is going to be optimized and what kind of weak learners will be used to approximate the solution. Continuous response problems, $y \in \mathbb{R}$, usually use either Quadratic loss, Absolute loss, Huber loss or Quantile loss, whereas binary response problems, $y \in  \{0,1\}$, typically apply Logistic loss or Adaboost loss. In this paper, we will consider Quadratic loss for continuous response and Logistic loss for binary response. To encourage diversity, we chose both decision trees and linear models as weak learners for GBMs.

When building a GBM, typically most time is spent on the selection of an optimal hyper-parameter. We automated this tedious tuning step of GBM by simply building various GBMs with different hyper-parameters and combined them using the stacked generalization framework. This multiple construction approach is inspired by random forests [5], of which are composed of a legion of independent decision trees used to form an ensemble. Calibrated Boosting-Forest utilizes this multiple construction technique and forms an ensemble from multiple GBMs each assigned a unique weight.

A key feature of CBF is the calibration of its output scoring. In some applications, and particularly in the drug discovery domain for the task of virtual screening  [7] where compounds are scored by the model based on likelihood of producing a drug-like effect in experimental testing, the mere ranking of examples (compounds) is not enough. Given the cost of testing each compound in high-throughput experimental screens, a measure of confidence in an estimated "hit rate" among some top ranking subset of compounds is an important factor in determining the number of compounds investigators should test experimentally in order to obtain some desired number of hits. In other words, how wide should the experimental net be cast given the anticipated abundance of "hits" based on model predictions. In this drug discovery application example as in many others, the output scores serve as inputs for subsequent procedures, such as estimating the cost/profit where the formula is $P(buy) * E(Pay \ amount | buy)$ or estimating total number of positive samples on test set where the formula is $\sum_{i=1}^{N} p_i$. The point being is that many applications rely on well calibrated prediction scores. And although GBMs built with decision tree base learners exhibit excellent ranking power, they consistently predict distorted probabilities [2]. Our Calibrated Boosting-Forest (CBF) model addresses this inadequacy without additional cost via altering the final model with a technique known as Platt scaling [12].

Calibrated Boosting-Forest has a notably significant value in the drug domain, specifically in regard to ligand-based virtual screening. In drug discovery, ligand-based virtual screening searches sets of molecules for desired biochemical activity typically producing continuous scores, such as a does- response curve or a normalized activity score based on the percentage change in molecule activity. After activity scoring, a domain expert then assigns a cutoff that determines which molecules are deemed active or inactive. Moreover, the particularly low hit/positive rate (lower than 1\%) and expensive cost to conduct real experiments makes calibrated probabilities a necessity. Because of the requisite well calibrated predictions, CBF can offer an improved method in this domain. Furthermore, beyond the merits of well calibrated predicted probability scores, our method, CBF, streamlines the tuning stage and provides equivalent or better ranking accuracy than the best individual GBMs.

\section{Methodology}
\label{headings}

\subsection{Extreme Gradient Boosting (XGBoost)}

For this study we chose the XGBoost [13] implementation of GBMs. XGBoost is a scalable boosting framework that supports building GBMs with weak learners based on both decision trees and linear models.
We use logistic loss as the loss function for the binary label, 
\begin{equation}
L(y_{n}, f(x_{n})) = -(y_{n} \times log(f(x_{n})) + (1-y_{n}) \times log(1-f(x_{n})))
\end{equation}
where $f(x_{n}) = p(y_{n} = 1 | x_{n}), f(x_{n}) \in (0,1)$\\
And quadratic loss for the continuous label,
\begin{equation}
L(y_{n}, f(x_{n})) = (f(x_{n}) - y_{n})^2
\end{equation}
where $f(x_{n}) = E(y_{n} | x_{n}), f(x_{n}) \in \mathbb{R}$\\
GBMs can be formalized in the additive form: $f(x) = \sum_{i=0}^{T}f_{i}(x)$ where $T$ is the number of iterations, $f_{0}$ is the initial weak learner, and $\{f_{i}\}_{i=1}^{T}$ are the additive weak learners.

\subsection{Calibrated Boosting-Forest Structure}

\subsubsection{Training}

Define $D^{train} = {(y_{n}, x_{n}), n = 1,...,N}$ where each $x_{n}$ is a feature vector and $y_{n}$ is the label of the $n_{th}$ instance and $M_{h}$ as a model where\\$h = \{(i,j,r), i \in {\{gbtree, gblinear\}}, j = hyper\;parameters, r = optimal\;round\}$\\
Gbtree and gblinear set the GBM to use decision trees or linear models, respectively, as the base learner. CBF randomly samples a unique hyper-parameters set for each GBM. The hyper-parameter space for gblinear is defined by lambda, alpha, lambda\_bias and learning rate. The hyper-parameter space for gbtree is defined by gamma, maximum depth, minimum child weight, maximum delta step, subsample ratio, column sample by tree ratio, column sample by level ratio, lambda, alpha and learning rate. During the k-fold cross-validation step, stratification splits the data $D^{train}$ into $K$ almost equal parts $D_{1},...,D_{k}$. Let $D_{k} = D - D_{k}$ denote the valid data and training data, respectively, for the $k_{th}$ fold. For each of the layer1 data, at each $k_{th}$ fold, $H$ models with different hyper-parameter sets $M_{1},...,M_{h}$ are learned from the training data $D^{(-k)}$ and these layer1 models $M_{1}^{(-k)},...,M_{h}^{(-k)}$ are applied to the valid data $D_{k}$. $H$ is a hyper-parameter selected to set the number of layer1 and layer2 models. Prediction scores of model $M_{h}^{(-k)}$ on data fold $D_{k}$ are denoted by
\begin{equation}
Z_{hkn} = M_{h}^{(-k)}(x_{n}), \text{ where } n \in D_{k}
\end{equation}
The concatenated predictions of all layer1 models, together with original label $y_{n}$ become the layer2 data, $MD_{k}=\{(y_{n}, Z_{1kn},...,Z_{hkn}),n \in D_{k}\}.$ 

During the cross-validation procedure, the only difference between distinct $M_{h}^{(-k)}$ with different $k$ is the number of optimal rounds/training steps, $r$. In other words, they all have same weak learner $i$ and parameter set $j$, but the optimal round $r$ is based on early stopping evaluated on $D_{k}$. After the cross-validation procedure, $MD=\cup_{k=1}^{K}MD_{k}$ becomes the full layer2 data with instance orders the same as $D$. Figure 1 illustrates the cross-validation procedure for one model. The layer2 model is trained using the same procedure as the layer1 models, with $x_{n}$ becoming the features from $MD$.
\begin{figure}[h!]
  \centering
  \includegraphics[scale=0.5]{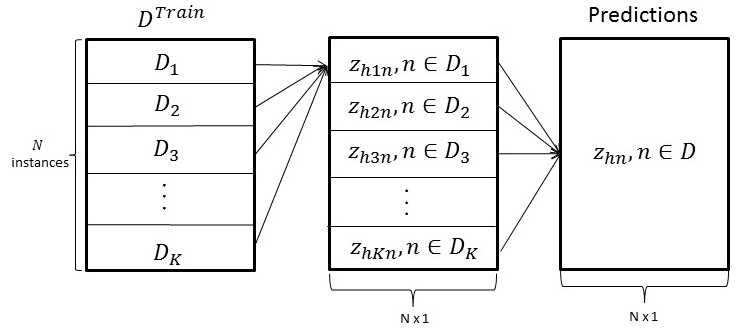}
  \caption{Training procedure of model $M_{h}$ }
\end{figure}

\subsubsection{Predicting}

When predicting on new data $D^{new}$, the learned layer1 models $M_{1}^{(-k)},...,M_{h}^{(-k)}$ all predict on the whole set of records of $D^{new}$ and produce $MD_{k}^{new}=\{(Z_{1kn},...,Z_{hkn}),n \in D^{new}\}$. Note that we have $K$ versions of $MD_{k}^{new}$ at this time, $MD^{new}=\frac{1}{K}\sum_{k=1}^{K}MD_{k}^{new}$ becomes the input features for the layer2 model. The layer2 model then uses the same procedure to generate the final predictions. Figure 2 illustrates the prediction procedure.
\begin{figure}[h!]
  \centering
  \includegraphics[scale=0.5]{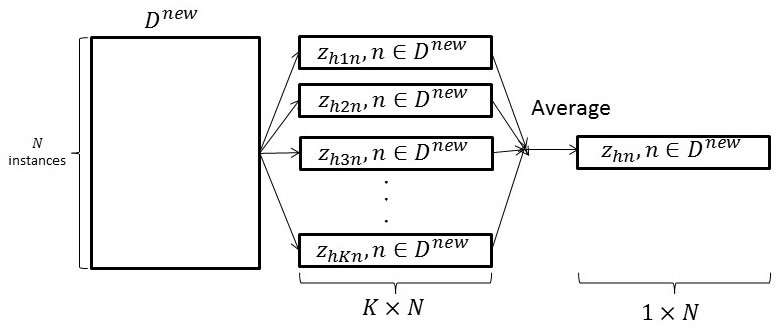}
  \caption{Predicting procedure of model $M_{h}$ }
\end{figure}

\subsubsection{Different labels}

In many situations, the raw labels follow a continuous distribution and a mapping function $f(x)$  then converts it into a binary label. In this paper, we denote the label from a continuous distribution as the continuous label and the label from a Bernoulli distribution as the binary label. The mapping function $f(x)$ can be based on a sophisticated relationship, or simply a threshold $t$ such that $f(x)=1, x>t \text{ and } f(x)=0, x \le t$. Under either decision, $f(x)$ is usually defined by a domain expert. When both label types are available, Calibrated Boosting-Forest builds two layer1 datasets, both using the same feature vectors where the first one uses a binary label and second one uses a continuous label. Each of the layer1 dataset will generate $H$ predictions, in total $2H$ predictions. When there is more than one layer1 dataset $D$, the final layer2 data is simply the horizontally concatenated version of multiple $MD$. For example, when we have both labels, we would have $D^{train, cont\;label}$ and $D^{train,binary\;label}$, which will generate $MD^{from\;cont\;label}$ and $MD^{from\;binary\;label}$. The concatenated versions of these two datasets, with binary label, become the final $MD$ that will be used to build layer2 model.

\subsubsection{Stopping criteria}

Since both base models, gbtree and gblinear are all based on gradient descent optimization, a proper stopping metric is needed in order to stop the model at the optimal number of iterations. We use one of the final evaluation metrics, Enrichment Factor, as the stopping metric. When building each model, we monitor the evaluation scores on training data $D^{(-k)}$ and valid data $D_{k}$. We stop training the model when we observe the evaluation score on $D_{k}$ does not improve for 100 rounds.The optimal number of rounds is the round that produces the best evaluation score on $D_{k}$.

When building the layer1 model based on continuous labels and the stopping metric requires a binary label to compute the score, we internally convert the continuous label into a binary label based on the same mapping function $f(x)$ discussed in section 3.2.3. Thus, the layer1 regression models are also built to optimize the same evaluation metric as the classification models.

\subsubsection{Platt scaling}

Platt scaling is a calibration method that passes the output from a single model through a sigmoid, resulting in output of posterior probabilities on [0,1]. This study extends the output mapping from a single model to multiple layer1 models and use $L_{1}$ and $L_{2}$-norms for regularization. To illustrate, suppose we have two sets of layer1 data, $D_{i},i=1,2$, each producing $H$ predictions. Under Platt scaling, these $2H$ layer1 predictions in addition to a column of 1s becomes the design matrix, \textbf{X},  in which we then train an elastic net [14] with \textbf{X} as the input feature and the binary label as the dependent variable. Denote $\beta \in \mathbb{R}^{2H+1}$ as the learned coefficients. The elastic net model is: $p(y=1|\textbf{X})=\frac{1}{1+\exp^{-X\beta}}$ where $\beta$ are found iteratively through gradient descent:
\begin{equation}
argmin_{\beta}\{-(y \times log(p)+(1-y)\times log(1-p) + \lambda_{2}|\beta|_{2}+\lambda_{1}|\beta|_{1}\}
\end{equation}
where $|\beta|_{2}=\sum_{j=1}^{2H+1}\beta_{j}^{2} \text{ and } |\beta|_{1} = \sum_{j=1}^{2H+1}|\beta_{j}|$.\\
The layer2 elastic net model performs two tasks: 
1. Learning the optimal weights for the combination of layer1 models in order to maximize ranking power
2. Performing probability/score calibration.
In section 4.3 we will introduce a novel metric for evaluating the quality of probability (Reliability score). Figure 3 illustrates the overall structure of Calibrated Boosting-Forest, with two layer1 data.
\begin{figure}[h!]
  \centering
  \includegraphics[scale=0.5]{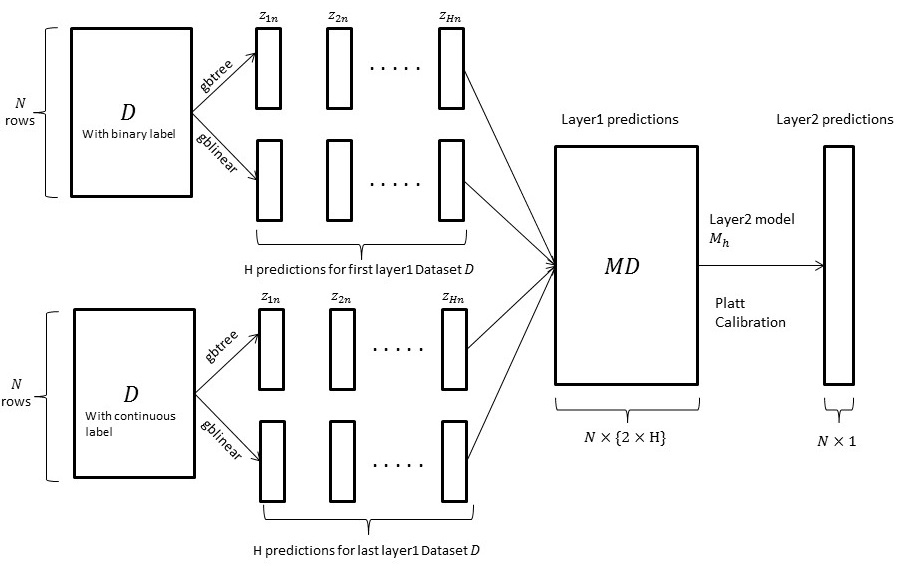}
  \caption{Calibrated Boosting-Forest structure }
\end{figure}

\subsubsection{Layer 2 model selection}
Mentioned in 3.2.1, there are H models for each layer1 $D_{i},i=1,2,...,N$ that together produce the final $MD$. While the training procedure of layer2 model (elastic net model) and layer1 (gbtree, gblinear) models is identical, the usage of $H$ predictions from $H$ layer2 models is different. Unlike layer1 models, whose predictions all contribute to layer2 feature data, the layer2 model will sweep all $H$ models and select the single model that has the best cross-validation scores. Notationally, we let $S$ be the evaluation metric, denote the valid score of $M_{h}^{(-k)}$ as: $CV_{hk}=S(Z_{hkn},y_{n}),\text{ where } n \in D_{k}$. And the cross-validation score of $M_{h}$ as: $CV_{h}=\frac{1}{K}\sum_{k=1}^{K}CV_{hk}$.

\section{Experimental design}
\subsection{Datasets}
Most of the existing machine learning models applied in virtual screening either directly model the binary label or the continuous label, but not both. Here we demonstrate Calibrated Boosting-Forest's utilization of diverse label types. We use the PCBA128 dataset from MoleculeNet [6] and add the original continuous label (logAC50) back to the dataset. PCBA128 contains 128 bioassays from PubChem BioAssay (PCBA), each bioassay represents an experimental screening of ~400,000 small molecules against some protein targets for some desired activity/effect. We consider one featurization method that is typical in ligand-based virtual screening applications, Extended Connectivity Fingerprint (ECFP) [11]. ECFP produces a fixed-length bit vector describing the atom connectivity in a small molecule.

\subsection{Data splitting}
In order to have the performance results comparable, we use the same random split from MoleculeNet. However, MoleculeNet has a different design of model structure compared to Calibrated Boosting-Forest. MoleculeNet's splitting methods all follow train/valid/test splits with default ratio 0.8/0.1/0.1. CBF has the same test split but a different train/valid splitting method that required further split the train/valid data into k folds, with each time having k-1 folds becoming the training data, 1 fold becoming valid data, and thus having k different validation scores. In order to make the results as comparable as possible, we keep the test and valid splits the same, and further split the train data into 4 folds, which gives us 5 train/valid folds. Besides reporting the overall train, valid, test performance of CBF, we also report the scores of same individual valid fold.

\subsection{Evaluation}
We use the same evaluation metric, AUC-PRC, as used by MoleculeNet. In addition, we add AUC-BED, AUC-ROC and Enrichment factors to cover different aspects of ranking performance and add Logistic loss and Reliability score to evaluate probability quality.

Area under the receiver operating characteristic (AUC-ROC) is a widely used metric for evaluating binary classification problems. ROC measures the true positive rate (TPR) vs. false positive rate (FPR) based on different thresholds. Area under the Boltzmann-Enhanced Discrimination (AUC-BED) [10] is similar to AUC-ROC whereas it puts greater emphasize on the early retrieval. This is more suitable to drug discovery since we want to limit testing to only the top scoring molecules. Area under the precision-recall curve (AUC-PRC) measures the positive predictive value and true positive rate over different thresholds. Enrichment factor (EF) at a given threshold t is another common used metric in virtual screening. The enrichment score reflects how many times better than random that a model's top t\% predictions are.

Logistic loss (Logloss) is the same one defined in section 2.1. When we don't know the true posterior probabilities, reliability diagrams [8] are used to visualize the relationship. [2] uses reliability diagrams to visualize the effectiveness of calibration on Adaboost. A reliability diagram is constructed by splitting the predictions into ten bins, with scores fall between 0 and 0.1 assigned to first bin, between 0.1 and 0.2 assigned to second bin, etc. For each bin, the mean predicted score and the true ratio of positive records are plotted against each other. We would expect the points fall near the diagonal line if the scores are well calibrated. However, when the ratio of positive labels is extremely low, lower than 1\% for example, the predictions from GBMs will have few records higher than 0.5. In this situation, the mean predicted value of higher bins becomes either highly fluctuated or N/A if no scores fall in that bin. To deal with this problem, we proposed an enhanced version that assigns 10\% of the records with lowest scores into first bin, 10\% of the records with lowest scores higher than previous one into second bin, etc. This procedure ensures that all the bins will have available scores and we would still expect the points fall near the diagonal line if the scores are well calibrated. Furthermore, to have a quantified and normalized score, we calculated the absolute difference between mean predicted value and true ratio of positive for each bin and divided by the true ratio of positive over the whole population and then take the arithmetic mean. We describe this value as a reliability score. The lower the reliability score is, the better the probabilities are calibrated.

\section{Results}
\subsection{Train/valid/test result of all models on ranking performance}
Models we tested include individual XGBoost and Calibrated Boosting-Forest with different number of H, denoted by XGBoost-H and CBF-H. Scores except XGBoost and CBF are from MoleculeNet [6]. XGBoost-H simply means we run H XGBoosts and choose the one with highest cross-validation score. MoleculeNet (MN) benchmark accesses the performances of various machine learning models, in particularly different neural network architectures, on a curated list of public experimental molecule screening datasets. It shows that graph convolutional model achieved a large performance boost because of the learnable featurizations. The train, valid and test scores report here are the average over all the targets. As we can see from table 1, individual GBM and CBF, regardless of which label used, all have scores higher than neural networks and logistic regression. Unless specific indication, the GBM and CBF reported all use both binary and continuous labels. Greater performance gains come from utilizing both labels type. CBF where H equals 1 and use both labels as input, has higher valid and test scores compared to best individual XGBoost, demonstrating the performance gain from utilizing diverse label types.

\begin{table}[h]
\centering
\caption{Performance between MoleculeNet and CBF on AUC-PRC (Mean)}
\label{my-label}
\begin{tabular}{@{}ccccc@{}}
\toprule
PCBA-128                & Train & Valid (DC's fold) & Valid (5 folds) & Test  \\ \midrule
Logistic Regression     & 0.166 & 0.130             & -               & 0.129 \\
Multitask Network       & 0.100 & 0.097             & -               & 0.100 \\
Bypass Network          & 0.121 & 0.111             & -               & 0.112 \\
Graph Convolution       & 0.151 & 0.136             & -               & 0.136 \\
CBF-1 (Binary)          & 0.450 & 0.177             & 0.147           & 0.163 \\
Best XGBoost-1 (Binary) & 0.480 & 0.155             & 0.148           & 0.165 \\
Best XGBoost-1          & 0.453 & 0.162             & 0.154           & 0.173 \\
CBF-1                   & 0.464 & \textbf{0.192}             & \textbf{0.181}           & \textbf{0.201} \\ \bottomrule
\end{tabular}
\end{table}

XGBoosts and CBFs reported here have large gaps between train and valid scores, indicating over-fitting is a general issue and proper regularization and stopping criteria are important. One possible explanation that models from MoleculeNet have smaller gap is that logistic regression and neural networks provided by MoleculeNet are multitasked, which serves as a strong regularization.

\subsection{Best individual XGBoost and Calibrated Boosting-Forest on probability quality performance}
Since Logloss and RS have different upper bounds depending on the distribution of response variables, we report median scores over 128 targets to control outliers. As we can see from table 2 and table 3, CBFs that incorporate Platt scaling achieve huge improvement in probability quality measurements. CBF with both label types achieved 0.82 on test reliability score, which means the average difference between estimated percentage of active samples and true percentage of active samples is around 82\% of actual percentage of active samples over all the bins. Having this property, we are able to quickly screen a large set of potential molecules and give a close estimate of the real active molecules over top K performing molecules. This can help users better estimate the cost and profit before doing real experiments. Although Logloss is hard to be directly translated in practical value, it is well known and we use it to demonstrate that the novel metric, RS, is working. CBF achieved improvement over both Logloss and RS.

\begin{table}[h]
\centering
\caption{Performance between best XGBoost and CBF on Logloss (Median)}
\label{my-label}
\begin{tabular}{@{}cccc@{}}
\toprule
PCBA-128              & Train & Valid         & Test         \\ \midrule
Best XGBoost-1 (Binary) & 32868 & 7444          & 3899         \\ \midrule
CBF-1 (Binary)        & 5152  & 1386          & 707          \\ \midrule
CBF-1                 & 3692  & \textbf{1007} & \textbf{559} \\ \bottomrule
\end{tabular}
\end{table}

\begin{table}[h]
\centering
\caption{Performance between best XGBoost and CBF on Reliability score (Median)}
\label{my-label}
\begin{tabular}{@{}cccc@{}}
\toprule
PCBA-128              & Train & Valid         & Test          \\ \midrule
Best XGBoost-1 (Binary) & 15.23 & 15.19         & 15.12         \\ \midrule
CBF-1 (Binary)        & 2.23  & 1.35          & 1.39          \\ \midrule
CBF-1                 & 1.36  & \textbf{0.74} & \textbf{0.82} \\ \bottomrule
\end{tabular}
\end{table}

\subsection{Scaling up CBF}
Table 4 demonstrates the effect of scaling up layer1 model with continuous label included, evaluated on various metrics. Since the computation will be long when applying to all 128 targets, we stratified sample 12 targets based on AUC-PRC performance of CBF-1 on PCBA128 test set. Thus these samples represent easy and challenging tasks from PCBA128. 

One thing notably is that performances of CBFs are all better than individual XGBoost. CBFs gain continuous improvement over all metrics, especially when increasing H from 5 to 10. Logloss and RS become similar for CBFs when H is greater than 1. Except for EF@0.01, best scores of rest evaluation metrics all come from CBFs with H equals 10 or 15. In practice, we would suggest choosing the highest H possible given the constraints of time and computation resource.

\begin{table}[h]
\centering
\caption{Effect of scaling up layer1 model (Test set)}
\label{my-label}
\begin{tabular}{@{}cccccccc@{}}
\toprule
PCBA-12         & AUC-ROC        & AUC-PRC        & AUC-BED        & EF@0.01        & EF@0.02        & Logloss         & RS            \\ \midrule
Best XGBoost-1  & 0.888          & 0.128          & 0.638          & 39.68          & 24.61          & 3828.51         & 22.46         \\ \midrule
Best XGBoost-5  & 0.885          & 0.131          & 0.634          & 37.62          & 23.26          & 3018.01         & 10.44         \\ \midrule
Best XGBoost-10 & 0.890          & 0.144          & 0.654          & 39.87          & 24.70          & 2183.90         & 8.55          \\ \midrule
Best XGBoost-15 & 0.890          & 0.141          & 0.649          & 39.42          & 24.67          & 1920.32         & 8.55          \\ \midrule
CBF-1           & 0.897          & 0.148          & 0.683          & \textbf{44.54} & 27.03          & 553.03          & 1.20          \\ \midrule
CBF-5           & 0.897          & 0.151          & 0.671          & 41.92          & 25.51          & 501.45          & 0.61          \\ \midrule
CBF-10          & \textbf{0.904} & 0.155          & \textbf{0.683} & 42.98          & 26.21          & 504.23          & \textbf{0.57} \\ \midrule
CBF-15          & 0.898          & \textbf{0.156} & 0.675          & 42.84          & \textbf{27.41} & \textbf{500.22} & 0.70          \\ \bottomrule
\end{tabular}
\end{table}

\section{Conclusion}
This work introduces Calibrated Boosting-Forest, an integrative method that offers advantageous characteristics to the deep-learning field. CBF minimizes the arduous hyper-parameter tuning step, handles both continuous and binary labels and yields well-calibrated posterior probabilities. Along with this new method, we introduce a novel probability evaluation metric that we coined Reliability Score. This metric offers an innovative way to measure and visualize probability calibration over rare event data by making use of quantiles. We demonstrate that CBFs can achieve superior ranking performance over state-of-the-art deep learning models while preserving well calibrated probabilities. As a general technique, we believe CBF can be applied advantageously to other applications in the future.

\subsubsection*{Acknowledgments}
The author thanks Anthony Gitter, Spencer Ericksen, and Liam Johnston who helped reviewing the paper and provided useful feedback. This work was funded by NIH/NCI P30 CA014520 to the UW Carbone Cancer Center and by the UW-Madison Office of the Vice Chancellor for Research and Graduate Education with funding from the Wisconsin Alumni Research Foundation.

\section*{References}
\small

[1] Wolpert, David H. "Stacked generalization." Neural networks 5, no. 2 (1992): 241-259.

[2] Niculescu-Mizil, Alexandru, and Rich Caruana. "Obtaining Calibrated Probabilities from Boosting." In UAI, p. 413. 2005. 

[3] Friedman, Jerome H. "Greedy function approximation: a gradient boosting machine." Annals of statistics (2001): 1189-1232.

[4] Breiman, Leo. "Stacked regressions." Machine learning 24, no. 1 (1996): 49-64.

[5] Breiman, Leo. "Random forests." Machine learning 45, no. 1 (2001): 5-32.

[6] Wu, Zhenqin, Bharath Ramsundar, Evan N. Feinberg, Joseph Gomes, Caleb Geniesse, Aneesh S. Pappu, Karl Leswing, and Vijay Pande. "MoleculeNet: A Benchmark for Molecular Machine Learning." arXiv preprint arXiv:1703.00564 (2017): v2.

[7] Geppert, Hanna, Martin Vogt, and Jurgen Bajorath. "Current trends in ligand-based virtual screening: molecular representations, data mining methods, new application areas, and performance evaluation." Journal of chemical information and modeling 50, no. 2 (2010): 205-216.

[8] DeGroot, Morris H., and Stephen E. Fienberg. "The comparison and evaluation of forecasters." The statistician (1983): 12-22.

[9] Zadrozny, Bianca, and Charles Elkan. "Obtaining calibrated probability estimates from decision trees and naive Bayesian classifiers." In ICML, vol. 1, pp. 609-616. 2001.

[10] Swamidass, S. Joshua, Chloe-Agathe Azencott, Kenny Daily, and Pierre Baldi. "A CROC stronger than ROC: measuring, visualizing and optimizing early retrieval." Bioinformatics 26, no. 10 (2010): 1348-1356.

[11] Rogers, David, and Mathew Hahn. "Extended-connectivity fingerprints." Journal of chemical information and modeling 50, no. 5 (2010): 742-754.

[12] Platt, John. "Probabilistic outputs for support vector machines and comparisons to regularized likelihood methods." Advances in large margin classifiers 10, no. 3 (1999): 61-74.

[13] Chen, Tianqi, and Carlos Guestrin. "Xgboost: A scalable tree boosting system." In Proceedings of the 22nd acm sigkdd international conference on knowledge discovery and data mining, pp. 785-794. ACM, 2016.

[14] Zou, Hui, and Trevor Hastie. "Regularization and variable selection via the elastic net." Journal of the Royal Statistical Society: Series B (Statistical Methodology) 67, no. 2 (2005): 301-320.

\end{document}